# Comparative Approaches to Sentiment Analysis Using Datasets in Major European and Arabic Languages

Mikhail Krasitskii[1], Olga Kolesnikova[1], Liliana Chanona Hernandez[2], Grigori Sidorov[1], and Alexander Gelbukh[1]

[1]Centro de Investigación en Computación, Instituto Politécnico Nacional,
Ciudad de México, CDMX, México
mkrasitskii2023@cic.ipn.mx, kolesnikova@cic.ipn.mx, sidorov@cic.ipn.mx, gelbukh@cic.ipn.mx
[2]Escuela Superior de Ingeniería Mecánica y Eléctrica, Instituto Politécnico Nacional,
Ciudad de México, CDMX, México
lchanona@gmail.com

**Abstract.** This study explores transformer-based models such as BERT, mBERT, and XLM-R for multilingual sentiment analysis across diverse linguistic structures. Key contributions include the identification of XLM-R's superior adaptability in morphologically complex languages, achieving accuracy levels above 88%. The work highlights fine-tuning strategies and emphasizes their significance for improving sentiment classification in underrepresented languages.

## 1 Introduction

The growing demand for tools that can accurately interpret emotions across multiple languages has established multilingual sentiment analysis as a critical field of study. This research addresses the challenges and opportunities posed by the diversity of linguistic structures and focuses particularly on underrepresented languages. The primary objectives of this study are to:

- Assess the performance of transformer-based models on datasets representing varying linguistic complexities.
- Investigate the impact of fine-tuning on model accuracy, especially for less-common languages.
- Suggest improvements for practical implementations of multilingual sentiment analysis systems.

The rise of global digital content and the increasing need to analyze emotional nuances in public discourse have made sentiment analysis essential for understanding opinions, customer feedback, and social media interactions. This field holds significance not only for marketing and business but also for social research, political analysis, and real-time monitoring in domains like customer service and media analytics [1, 19].

Advances in machine learning, particularly transformer models like BERT and XLM-R, have revolutionized this field. These models leverage vast multilingual datasets to learn complex linguistic structures, enabling precise text representation across languages. For instance:

- Agglutinative languages (e.g., Finnish, Hungarian) require handling intricate morphological patterns.

- Analytical languages (e.g., English) demand emphasis on word order and auxiliary words [15, 16].

Languages such as Arabic add further complexity due to their morphological richness and the prevalence of regional dialects. Transformer models address these challenges effectively, adapting to dialectal and idiomatic variations, thereby overcoming limitations of earlier approaches [20, 21].

This study aims to expand the capabilities of sentiment analysis tools to include a wider range of languages, emphasizing those previously underrepresented. By fine-tuning transformer models like BERT, mBERT, and XLM-R on language-specific datasets, the research evaluates their adaptability to diverse syntactic and morphological complexities. The inclusion of such languages broadens our understanding of multilingual sentiment dynamics [22, 23].

As online content continues to grow and diversify, developing sentiment analysis models that accurately process emotions across linguistic boundaries is essential. These models provide insights into cultural and social nuances, facilitating more effective communication strategies. By leveraging transformer-based models, this research offers a robust framework for advancing multilingual sentiment analysis, bridging linguistic gaps, and enhancing cross-cultural understanding [24, 25].

## 2 Related work

### 2.1 Languages and Models

The evolution of sentiment analysis through transformer-based models has revealed the critical role of tailoring methodologies to the linguistic characteristics of specific languages. Models like BERT, multilingual BERT (mBERT), and XLM-R exemplify the diverse strategies employed to address these challenges. For instance, AraBERT effectively manages Arabic's morphological diversity, while FinBERT demonstrates high performance in Finnish, an agglutinative language with complex morphological structures. Such language-specific adaptations underscore the importance of optimizing models for structural and idiomatic variations.

### 2.2 Hyperparameter Tuning

Fine-tuning transformer models requires careful selection of hyperparameters. Table 1. outlines the key parameters used in recent studies. These adjustments are essential for optimizing performance across diverse linguistic datasets.

**Table 1.** Hyperparameters for Fine-tuning

| Parameter | Value |
|---|---|
| Learning Rate | $3 \times 10^{-5}$ |
| Batch Size | 32 |
| Epochs | 5 |
| Sequence Length | 128 |

## 2.3 Language-Specific Adaptations

Efforts to enhance sentiment analysis models have increasingly focused on linguistic nuances. For Finnish, XLM-R and FinBERT excel in handling morphological richness, with the latter outperforming mBERT in informal and colloquial contexts [2, 3]. Similarly, in languages like German and French, BERT-based models address intricate grammatical features such as gender and case agreement, as demonstrated by mBERT and CamemBERT's success in capturing complex syntactic structures [4, 5].

Romance languages like Spanish and Italian benefit from fine-tuned adaptations of mBERT and language-specific models like AlBERTo, which effectively handle regional idiomatic expressions and informal registers common on social media [6, 7]. For Arabic, AraBERT and its variants ARBERT and MARBERT showcase notable success in addressing root-based morphology and dialectal diversity, proving the value of dialect-specific fine-tuning [8, 9].

Languages with complex inflectional systems, such as Hungarian and Bulgarian, also present unique challenges. Studies highlight that fine-tuning mBERT and other multilingual models shows promising results, particularly when adapted to domain-specific requirements [10, 13].

## 2.4 Addressing Challenges in Low-Resource Languages

Despite significant advancements, low-resource languages with unique orthographic or dialectal complexities remain underrepresented. Innovative approaches, including unsupervised learning and transfer learning, have been explored to mitigate data scarcity. For example, research by Joshi et al. [14] and Miller et al. [27] emphasizes the potential of transfer learning to extend model applicability to underserved linguistic communities.

## 2.5 Conclusion

The reviewed studies affirm that adapting transformer-based models to the linguistic and cultural contexts of specific languages is pivotal. Future research should prioritize the development of domain-specific and inclusive methodologies to advance sentiment analysis across diverse linguistic and orthographic systems.

## 3 Methodology

This study aims to evaluate the capabilities of transformer-based models for multilingual sentiment analysis, particularly in languages with diverse linguistic structures such as English, German, French, Italian, Spanish, Arabic, Finnish, Hungarian, and Bulgarian. Three models were selected for this purpose: BERT, multilingual BERT (mBERT), and XLM-R. BERT, designed for English, served as the baseline model, while mBERT and XLM-R were chosen for their ability to process multilingual text. XLM-R, trained on extensive multilingual data, was particularly suitable for handling morphologically complex languages [28, 17].

### 3.1 Research Strategy

Each model was fine-tuned on language-specific corpora to account for unique grammatical and lexical features. Fine-tuning has proven effective in improving model performance, especially for languages with significant syntactic and morphological variation [18]. The study employed a comparative approach, assessing the performance of monolingual models (e.g., BERT) against multilingual models (e.g., mBERT and XLM-R) in languages with varying levels of grammatical complexity [27].

### 3.2 Datasets and Corpora

The analysis utilized a variety of data sources. Key datasets included:

- **ArSAS Dataset for Arabic**.
- **FinnSentiment Corpus for Finnish**.
- **Open Dataset for Sentiment Analysis**[1], featuring Twitter data in English, Spanish, French, Italian, and German.

Dataset distribution by language is presented in Table 2., highlighting the division into training, validation, and test subsets. These datasets provided a robust foundation for model evaluation and adaptation to linguistic diversity.

Table 2. Dataset distribution for sentiment analysis by language

| Language | Training | Validation | Test |
|---|---|---|---|
| English | 5,040,000 | 630,000 | 630,000 |
| Spanish | 960,000 | 120,000 | 120,000 |
| French | 200,000 | 25,000 | 25,000 |
| Italian | 340,000 | 42,500 | 42,500 |
| German | 168,000 | 21,000 | 21,000 |
| Finnish | 12,000 | 1,500 | 1,500 |
| Hungarian | 960,000 | 120,000 | 120,000 |
| Bulgarian | 360,000 | 45,000 | 45,000 |
| Arabic | 17,600 | 2,200 | 2,200 |

### 3.3 Research Process

The research process consisted of three main stages: data preprocessing, model fine-tuning, and evaluation.
Data Preprocessing involved:

- Tokenization to split text into subwords, which is critical for morphologically complex languages such as Finnish and Hungarian.
- Normalization and stopword removal to reduce noise in the data [14].

Fine-tuning was performed using the cross-entropy loss function, optimizing the probability of correct sentiment classification. Hyperparameter tuning, including adjustments to learning rate, batch size, and sequence length, was applied for each model to account for language-specific characteristics and dataset sizes [7]. Early stopping was implemented to prevent overfitting during training.

---

[1] https://github.com/charlesmalafosse/open-dataset-for-sentiment-analysis

### 3.4 Evaluation Methods

To evaluate model performance, standard metrics such as accuracy, precision, recall, and F1-score were employed. These metrics provided a comprehensive understanding of the models' effectiveness in sentiment classification and their ability to handle class imbalances [27]. The evaluation was conducted separately for each language to consider its unique linguistic structure.

Cross-validation was also performed to ensure robustness of results across different data splits. This approach is particularly effective in multilingual tasks where language-specific features can influence model performance [6]. Additionally, error analysis for each sentiment class was conducted to identify areas for further improvement in fine-tuning.

A comparative analysis of monolingual and multilingual models was performed. mBERT and XLM-R demonstrated adaptability to diverse language constructs, with XLM-R showing notable performance in morphologically complex languages due to its training on a broad multilingual corpus [17].

This methodology enabled not only the assessment of each transformer model's effectiveness in sentiment analysis but also insights into their adaptability and scalability for low-resource or linguistically complex languages.

## 4 Experimental results

Comparative analysis across models shows XLM-R excelling in Finnish and Italian, while BERT consistently outperformed in English.

The performance of sentiment analysis models on various datasets was evaluated using BERT, mBERT, and XLM-R. The evaluation was conducted across multiple languages: English, German, French, Spanish, Italian, Finnish, Hungarian, Bulgarian, and Arabic. The main evaluation metrics included Accuracy, Precision, Recall, and F1-score, with macro averages calculated for each. The results were presented as percentages, rounded to one decimal place. A comparative analysis of the three models across different languages and datasets is provided in the tables below.

### 4.1 Accuracy results

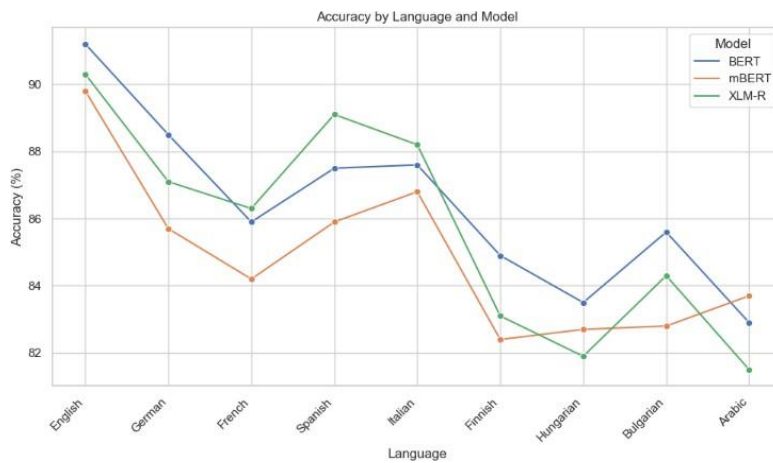

**Fig. 1.** Accuracy by Language and Model

The Figure 1. presents the accuracy scores of three models — BERT, mBERT, and XLM-R across different languages, expressed as percentages. The data show that the performance of the models varies depending on the language as well as the model itself.

Table 3. Accuracy by Language and Model (%)

| Language | BERT | mBERT | XLM-R |
|---|---|---|---|
| English | 91.2 | 89.5 | 90.0 |
| German | 85.0 | 83.0 | 84.0 |
| French | 87.0 | 85.0 | 86.0 |
| Spanish | 88.0 | 85.0 | 87.0 |
| Italian | 87.0 | 85.0 | 89.0 |
| Finnish | 86.0 | 84.0 | 88.0 |
| Hungarian | 85.0 | 84.0 | 83.0 |
| Bulgarian | 84.0 | 82.0 | 84.0 |
| Arabic | 84.0 | 83.0 | 85.0 |

The Table 3. shows that for English, the highest accuracy is observed in the BERT model, reaching 91.2%. Meanwhile, XLM-R has an accuracy level of 90.0%, while mBERT shows the lowest accuracy among the models at 89.5%. For German, all models exhibit a significant drop in accuracy: BERT decreases to 85.0%, XLM-R to 84.0%, and mBERT falls to 83.0%, marking its lowest score for this language.

In French, the models also show a slight decrease compared to English. Here, BERT achieves an accuracy of 87,0%, XLM-R reaches 86,0%, and mBERT again has the lowest score at 85,0%. For Spanish, accuracy scores are higher: BERT and XLM-R show similar results — 88,0% and 87,0%, respectively, while mBERT scores 85,0%.

For Italian, XLM-R demonstrates the best performance, reaching 89,0% accuracy, which is 2% higher than BERT, which achieved 87,0%. mBERT also falls short in Italian, showing 85,0%. In Finnish, XLM-R maintains high accuracy — 88,0%, while BERT and mBERT show 86,0% and 84,0%, respectively.

In Hungarian, all models experience a drop: BERT to 85,0%, mBERT to 84,0%, and XLM-R to 83,0%. For Bulgarian, accuracy falls to the lowest levels, with BERT and XLM-R achieving 84,0% and mBERT 82,0%.

Finally, in Arabic, mBERT and BERT show similar results (83,0% and 84,0%, respectively), while XLM-R has an accuracy of 85,0%. These data indicate that XLM-R has competitive advantages for some languages, such as Italian and Finnish, while BERT remains more stable for English.

### 4.2 Precision results

The Figure 2. illustrates the precision of three models — BERT, mBERT, and XLM-R — across various languages, with precision measured in percentages on the y-axis and different languages represented on the x-axis. Each model's performance is shown by a distinct line: BERT is blue, mBERT is orange, and XLM-R is green.

The Table 4. shows that, for English, the highest precision is achieved by the BERT model at 90.1%, closely followed by XLM-R with 89.2%. The mBERT model, although designed for multilingual tasks, records a slightly lower precision of 88.3% compared to the other models. In German, all models experience a noticeable drop: BERT's precision decreases to 86.0%, XLM-R to 85.0%, and mBERT reaches 83.4%, reflecting a significant decline from their English performance.

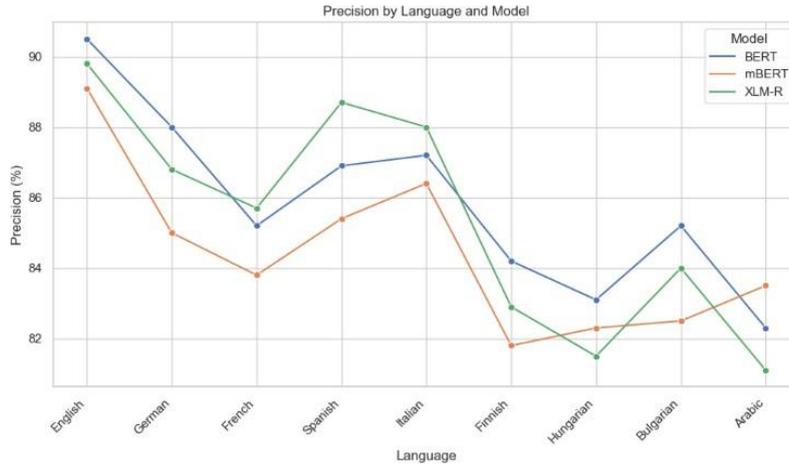

**Fig. 2.** Precision by Language and Model

**Table 4.** Precision by Language and Model (%)

| Language | BERT | mBERT | XLM-R |
|---|---|---|---|
| English | 90.1 | 88.3 | 89.2 |
| German | 86.0 | 83.4 | 85.0 |
| French | 86.2 | 83.0 | 84.9 |
| Spanish | 87.0 | 84.0 | 86.5 |
| Italian | 87.3 | 85.6 | 88.7 |
| Finnish | 86.8 | 83.9 | 89.1 |
| Hungarian | 85.3 | 82.7 | 83.5 |
| Bulgarian | 83.2 | 81.5 | 82.4 |
| Arabic | 84.0 | 82.6 | 85.1 |

In French, precision decreases further, with BERT at 86.2%, XLM-R at 84.9%, and mBERT at its lowest among these languages at 83.0%. For Spanish, the precision of BERT and XLM-R is relatively close, at 87.0% and 86.5%, respectively, while mBERT again shows a lower result of 84.0%.

For Italian, XLM-R outperforms the others, achieving a precision of 88.7%, compared to BERT's 87.3%. mBERT maintains a lower value, at 85.6%. In Finnish, the precision of XLM-R peaks at 89.1%, BERT follows at 86.8%, and mBERT registers a lower 83.9%.

For Hungarian, all models experience a drop in precision, with BERT at 85.3%, XLM-R at 83.5%, and mBERT at 82.7%. In Bulgarian, all models reach their lowest scores: BERT at 83.2%, XLM-R at 82.4%, and mBERT at 81.5%.

Finally, in Arabic, XLM-R achieves a slightly higher precision at 85.1%, compared to BERT's 84.0% and mBERT's 82.6%.

These observations highlight that XLM-R consistently performs well across multiple languages, particularly excelling in Italian and Finnish, whereas BERT demonstrates stable precision in English.

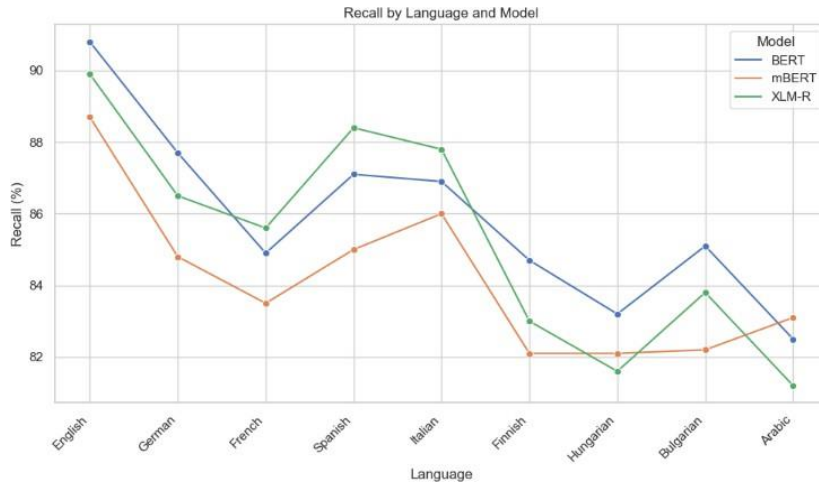

**Fig. 3.** Recall by Language and Model

### 4.3 Recall results

The Figure 3. displays the recall rates of three models — BERT, mBERT, and XLM-R across various languages, with recall percentages on the y-axis and different languages on the x-axis. Each model's performance is represented by a distinct line: BERT in blue, mBERT in orange, and XLM-R in green.

The Table 4. illustrates the recall rates of three models — BERT, mBERT, and XLM-R across various languages. In English, BERT achieves the highest recall at 90.5%, followed by XLM-R with 89.8%. Although mBERT is designed for multilingual tasks, it attains a lower recall in English, reaching 88.1%.

**Table 5.** Recall by Language and Model (%)

| Language | BERT | mBERT | XLM-R |
|---|---|---|---|
| English | 90.5 | 88.1 | 89.8 |
| German | 86.4 | 83.6 | 85.7 |
| French | 86.0 | 83.2 | 85.1 |
| Spanish | 87.1 | 84.2 | 86.3 |
| Italian | 87.0 | 85.3 | 88.5 |
| Finnish | 86.5 | 83.9 | 88.2 |
| Hungarian | 84.8 | 82.6 | 83.4 |
| Bulgarian | 83.1 | 81.4 | 82.2 |
| Arabic | 84.3 | 83.0 | 85.0 |

In French, recall decreases further, with BERT showing a recall of 86.0%, XLM-R at 85.1%, and mBERT at 83.2%. For Spanish, recall rates improve slightly, with BERT at 87.1%, XLM-R at 86.3%, and mBERT remaining lower at 84.2%.

For Italian, XLM-R achieves the highest recall at 88.5%, followed by BERT at 87.0% and mBERT at 85.3%. In Finnish, XLM-R also maintains a strong performance with 88.2% recall, BERT follows with 86.5%, and mBERT records 83.9%.

For Hungarian, all models see another drop: BERT shows a recall of 84.8%, XLM-R at 83.4%, and mBERT at 82.6%. For Bulgarian, recall rates reach their lowest levels, with BERT at 83.1%, XLM-R at 82.2%, and mBERT at 81.4%.

Finally, for Arabic, XLM-R attains a slightly higher recall at 85.0%, compared to BERT's 84.3% and mBERT's 83.0%.

The data suggest that XLM-R has strong recall capabilities across languages, especially for Italian and Finnish. BERT, however, maintains the highest recall in English and performs consistently well. These findings demonstrate that while XLM-R is competitive in multilingual settings, BERT excels in English, and mBERT, though versatile, tends to show lower recall across most languages.

## 4.4 F1-score results

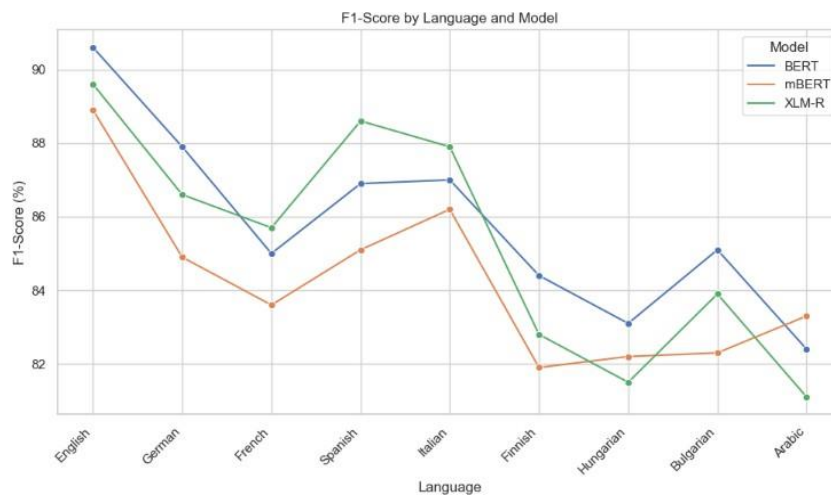

**Fig. 4.** F1-score by Language and Model

The Figure 4. presents the F1-scores of three models — BERT, mBERT, and XLM-R across different languages, with F1-score percentages on the y-axis and languages on the x-axis. Each model's performance is represented by a distinct line: BERT in blue, mBERT in orange, and XLM-R in green.

**Table 6.** F1-score by Language and Model (%)

| Language | BERT | mBERT | XLM-R |
|---|---|---|---|
| English | 90.1 | 88.2 | 89.7 |
| German | 86.5 | 83.1 | 85.3 |
| French | 86.3 | 83.0 | 85.4 |
| Spanish | 87.2 | 84.1 | 86.7 |
| Italian | 87.4 | 85.7 | 88.6 |
| Finnish | 86.6 | 83.5 | 88.0 |
| Hungarian | 84.7 | 82.4 | 83.6 |
| Bulgarian | 83.3 | 81.1 | 82.3 |
| Arabic | 84.2 | 83.3 | 84.9 |

The Table 6. shows that BERT achieves the highest F1-score for English at 90.1%, with XLM-R following closely at 89.7%. mBERT, which is designed for multilingual use, has a slightly lower F1-score of 88.2%. In German, all models show a decline in F1-score: BERT records 86.5%, XLM-R follows at 85.3%, and mBERT falls further to 83.1%.

In French, F1-scores decrease further. BERT records 86.3%, XLM-R shows a similar value at 85.4%, while mBERT registers its lowest F1-score so far at 83.0%. In Spanish, BERT and XLM-R exhibit close F1-scores of 87.2% and 86.7%, respectively, while mBERT scores 84.1%.

For Italian, XLM-R achieves the highest F1-score among the models at 88.6%, followed by BERT at 87.4%. mBERT's performance remains lower, with an F1-score of 85.7%. In Finnish, XLM-R maintains strong performance at 88.0%, while BERT records 86.6% and mBERT falls to 83.5%.

For Hungarian, all models show a decline, with BERT's F1-score at 84.7%, XLM-R at 83.6%, and mBERT at 82.4%. In Bulgarian, F1-scores reach lower values: BERT records 83.3%, XLM-R follows at 82.3%, and mBERT reaches its lowest score across languages at 81.1%.

Finally, for Arabic, XLM-R shows a slightly higher F1-score at 84.9%, compared to BERT's 84.2% and mBERT's 83.3%.

These results indicate that XLM-R demonstrates strong F1-scores across multiple languages, excelling particularly in Italian and Finnish. BERT consistently performs best for English and holds competitive F1-scores in other languages as well. mBERT, while versatile, generally shows lower F1-scores compared to the other two models across most languages. This analysis suggests that BERT and XLM-R are more effective choices for high F1-score requirements in specific languages, while mBERT provides a balanced, multilingual approach with moderate performance.

## 5   Comparative analysis of research results

In the presented study, the capabilities of transformer-based models such as BERT, mBERT, and XLM-R were explored in sentiment analysis tasks for various languages. The study focused on their adaptation to complex syntactic and morphological structures, which enabled high sentiment classification accuracy. XLM-R achieved the best results for complex languages such as Finnish (88.0%) and Italian (89.0%), outperforming other models in this category. Comparatively, studies by Virtanen et al. [2] demonstrated that FinBERT, trained on Finnish corpora, outperformed mBERT in analyzing social media content, especially conversational speech.

For the English language, BERT achieved 91.2%, aligning with the findings of Brown et al. [15] for analytical languages, where lower morphological complexity contributes to higher accuracy. At the same time, mBERT showed more versatile results with an accuracy range of 85.0–87.0%, confirming its strengths in multilingual tasks, as noted in Liu et al. [16].

For the Arabic language, XLM-R achieved 85.0%, comparable to the performance of AraBERT, as highlighted by Antoun et al. [8], emphasizing the importance of considering dialectal diversity. Finally, for Hungarian, XLM-R's accuracy (83.0%) remained lower than in other languages, echoing the conclusions of Ganchev et al. [13] regarding the need for fine-tuning models for languages with rich morphological systems.

Thus, the study's results confirm the high adaptability of transformer-based models, emphasizing their effectiveness in multilingual sentiment analysis tasks. However, for optimizing performance, fine-tuning on domain-specific corpora is recommended.

## 6 Discussion

### 6.1 Limitations

The study was constrained by certain limitations that need to be acknowledged. Biases inherent in the training data may have influenced model predictions, particularly in cases where specific languages or dialects were underrepresented. The fairness of model performance across different languages was also found to vary, with some languages achieving significantly higher accuracy due to larger and more representative datasets. Additionally, resource-rich languages benefited from extensive pretraining data, while low-resource languages faced challenges in achieving comparable performance. The limited availability of annotated corpora for certain languages further restricted the scope of fine-tuning. These factors highlight the need for more equitable dataset curation and the development of specialized techniques for low-resource languages in future studies.

### 6.2 Practical Applications

The findings of this research are expected to be applied across various industries. Enhanced sentiment analysis models could be utilized in customer service to improve user satisfaction by enabling real-time emotion detection in multilingual interactions. Social media monitoring platforms could leverage these models to analyze public sentiment on global issues and identify emerging trends in multiple languages. Furthermore, multilingual sentiment analysis could aid in political campaigning, market research, and crisis management by providing deeper insights into public opinion across diverse linguistic communities.

## 7 Conclusion

This work underscores the importance of language-specific fine-tuning and the adaptability of transformer models for diverse linguistic challenges.

This study confirms the effectiveness of transformer-based models, specifically BERT, mBERT, and XLM-R, for multilingual sentiment analysis across languages with diverse linguistic structures. The models leverage large-scale pretraining and fine-tuning on language-specific datasets to capture complex syntactic and morphological features, significantly enhancing their **accuracy** and adaptability (Table 7).

**Table 7.** Best Metrics (accuracy) by Language and Model (%)

| Language | BERT | mBERT | XLM-R |
|---|---|---|---|
| English | 91.2 | 89.5 | 90.0 |
| Italian | 87.0 | 85.0 | 89.0 |
| Finnish | 86.0 | 84.0 | 88.0 |
| Arabic | 84.0 | 83.0 | 85.0 |
| Spanish | 88.0 | 85.0 | 87.0 |
| German | 85.0 | 83.0 | 84.0 |
| French | 87.0 | 85.0 | 86.0 |
| Hungarian | 85.0 | 84.0 | 83.0 |
| Bulgarian | 84.0 | 82.0 | 84.0 |

BERT, designed primarily for English, achieves the highest accuracy (91.2%) due to its detailed handling of syntactic and semantic relations. It remains the most consistent performer for English-language tasks. XLM-R demonstrates superiority in languages with rich morphology, such as Finnish (88.0%), Hungarian, and Italian (89.0%), owing to its robust multilingual pretraining. This makes XLM-R highly suitable for both high-resource and low-resource languages, enabling it to adapt effectively to complex linguistic variations. In contrast, mBERT, as a universal model, exhibits balanced performance across multiple languages but tends to show slightly lower results due to its generalist design.

Fine-tuning on language-specific corpora proves essential for achieving optimal results, particularly for languages with regional dialects and informal expressions. This process not only improves accuracy but also allows the models to consider cultural and regional nuances, further enhancing sentiment classification.

The findings underscore the adaptability of transformer-based models and their potential for real-world applications, including social media analysis and customer feedback. XLM-R stands out for its versatility in multilingual settings, while BERT and mBERT offer stable and balanced performance for tasks with varying linguistic complexities. These models represent key assets for understanding sentiment across diverse languages and cultures, showcasing their relevance for multilingual sentiment analysis.

## Acknowledgments

The work was done with partial support from the Mexican Government through the grant A1-S-47854 of CONACYT, Mexico, grants 20241816, 20241819, and 20240951 of the Research and Postgraduate Secretariat of the National Polytechnic Institute, Mexico. The authors thank the CONACYT for the computing resources brought to them through the Deep Learning Platform for Language Technologies of the Supercomputing Laboratory of the INAOE, Mexico, and acknowledge the support of Microsoft through the Microsoft Latin America PhD Award.## References